\pdfoutput=1

\documentclass[11pt]{article}

\usepackage{acl}

\usepackage{times}
\usepackage{latexsym}

\usepackage[T1]{fontenc}

\usepackage[utf8]{inputenc}

\usepackage{microtype}

\usepackage{makecell}

\usepackage{pifont}
\newcommand{\cmark}{\ding{51}}
\newcommand{\xmark}{\ding{55}}

\usepackage{color}

\newcommand{\forcamera}[1]{}
\usepackage[disable]{todonotes}
\usepackage{float}

\newcommand{\caancora}[0]{\textsf{ca\_ancora}}
\newcommand{\cspcedt}[0]{\textsf{cs\_pcedt}}
\newcommand{\cspdt}[0]{\textsf{cs\_pdt}}
\newcommand{\engum}[0]{\textsf{en\_gum}}
\newcommand{\enparcorfull}[0]{\textsf{en\_parcorfull}}
\newcommand{\frdemocrat}[0]{\textsf{fr\_democrat}}
\newcommand{\deparcorfull}[0]{\textsf{de\_parcorfull}}
\newcommand{\depotsdamcc}[0]{\textsf{de\_potsdam}}
\newcommand{\huszegedkoref}[0]{\textsf{hu\_szeged}}
\newcommand{\ltlcc}[0]{\textsf{lt\_lcc}}
\newcommand{\plpcc}[0]{\textsf{pl\_pcc}}
\newcommand{\rurucor}[0]{\textsf{ru\_rucor}}
\newcommand{\esancora}[0]{\textsf{es\_ancora}}

\newcommand{\baseline}[0]{\textsc{Baseline}} 

\title{Findings of the Shared Task on Multilingual Coreference Resolution}

\author{
Zdeněk Žabokrtský$^1$,
Miloslav Konopík$^2$,
Anna Nedoluzhko$^1$,
Michal Novák$^1$, \\
\textbf{Maciej Ogrodniczuk$^3$,
Martin Popel$^1$,
Ondřej Pražák$^2$,} \\
\textbf{Jakub Sido$^2$,
Daniel Zeman$^1$,
Yilun Zhu$^4$} \\[2mm]
$^1$ Charles University, Faculty of Mathematics and Physics, \\ Institute of Formal and Applied Linguistics, Prague, Czechia \\
\texttt{\{zabokrtsky,nedoluzhko,mnovak,popel,zeman\}@ufal.mff.cuni.cz}\\[2mm]
$^2$
University of West Bohemia, Faculty of Applied Sciences, \\ Department of Computer Science and Engineering, Pilsen, Czechia \\
\texttt{konopik@kiv.zcu.cz, \{ondfa,sidoj\}@ntis.zcu.cz}\\[2mm]
$^3$
Institute of Computer Science, Polish Academy of Sciences \\
Warsaw, Poland, \texttt{maciej.ogrodniczuk@gmail.com}\\[2mm]
$^4$ Georgetown University, Department of Linguistics, \\ Washington, DC, USA, \texttt{yz565@georgetown.edu}
}

\usepackage{booktabs}

\newcommand{\ndatasets}[0]{13}
\newcommand{\nlanguages}[0]{10}
\newcommand{\nsystems}[0]{8}
\newcommand{\nteams}[0]{5}

\begin{document}
\maketitle

\begin{abstract}
This paper presents an overview of the shared task on multilingual
coreference resolution associated with the CRAC 2022 workshop. Shared task
participants were supposed to develop trainable systems capable of
identifying mentions and clustering them according to identity
coreference. The public edition of CorefUD~1.0, which contains \ndatasets{}
datasets for \nlanguages{} languages, was used as the source of training and
evaluation data. The CoNLL score used in previous coreference-oriented
shared tasks was used as the main evaluation metric. There were \nsystems{}
coreference prediction systems submitted by \nteams{} participating teams;
in addition, there was a competitive Transformer-based baseline system
provided by the organizers at the beginning of the shared task. The winner
system outperformed the baseline by 12 percentage points (in terms of the
CoNLL scores averaged across all datasets for individual languages).
\end{abstract}

\section{Introduction}

Multilingual shared tasks are an important source of momentum in various
subfields of NLP research, with the CoNLL-X shared task
on multilingual dependency parsing \cite{buchholz2006conll} being one of the
most successful and influential examples. Clearly, the limiting factor for
organizing such shared tasks is the availability of multilingual data whose
annotations are harmonized at least to some extent, so that the experiments
on individual languages can be performed and evaluated in a uniform way.

In the coreference world, one of the first multilingual shared tasks were
SemEval-2010 \citep{recasens-etal-2010-semeval} with seven languages and
CoNLL-2012 \cite{conll-2012}, in which OntoNotes data for three languages (English,
Chinese, and Arabic) were included.
With the recent advance of the CorefUD collection
\cite{ourtechrep2021, corefud2022lrec}, harmonized coreference data for \nlanguages{} languages (covered in CorefUD's publicly available edition) became available. Hence, CorefUD is the source of data for the
present shared task; more information about the collection is given in
Section~\ref{sec:data}. In brief, participants of this shared task are
supposed to (a) identify mentions in texts and (b) predict which mentions
belong to the same coreference cluster (i.e., refer to the same entity or
event), using the CorefUD data both for training and evaluation of their
coreference resolution systems.

A specific feature of CorefUD is that it combines coreference with dependency
syntax, using the annotation scheme (and file format too) of the Universal
Dependencies (UD) project \cite{ud}. In all datasets included in the collection,
the coreference annotation is manual and the dependency annotation is either
manual too, if available, or produced by a dependency parser. Empirical
evidence showing advantages of such symbiosis of coreference and dependency
syntax is presented in two case studies   \cite{2021findings,
onehead-depling}. Participants of this shared task can employ the dependency
annotation for determining mention spans (as mentions often correspond to
syntactically meaningful units) and for determining core parts of mentions
(which correspond to syntactic head in CorefUD).

To the best of our knowledge, this is the first shared task on multilingual coreference resolution that accepts zeros (e.g. elided subjects) as potential members of coreference chains.%
\footnote{%
\citet{recasens-etal-2010-semeval} do not state how zeros were treated for pro-drop languages such as Spanish and Catalan in SemEval-2010, and \citet{conll-2012} excluded all zeros from the CoNLL-2012 shared task data.
}
Zeros are an integral part of some of the CorefUD datasets, using empty nodes in enhanced UD representation to annotate them.
We keep all annotated zeros, encouraging participants to resolve coreference also for this type of potential mentions.

As with other shared tasks, evaluation is crucial.
Unfortunately, and unlike e.g. in dependency parsing, there is no simple and
easily interpretable accuracy metric for coreference. We adhere to using the
CoNLL score developed in former coreference shared tasks. More specifically,
we use an average of the $F_1$ values of MUC, B$^3$ and CEAF-e scores as the
main evaluation metric. More details concerning evaluation are presented in
Section~\ref{sec:evaluation}.

A Transformer-based coreference prediction system
\cite{pravzak2021multilingual} was provided as a strong baseline to the shared
task participants. The baseline system as well as \nsystems{} systems
submitted by the participants are briefly described in
Section~\ref{sec:systems} and some of the systems are described in more
detail in separate papers in this volume.
Their results are summarized in
Section~\ref{sec:results}.
Possible directions for future editions of the shared task are outlined in
Section~\ref{sec:conclusions}.

\section{Datasets}
\label{sec:data}

For training and evaluation purposes, the present shared task uses
\ndatasets{} coreference datasets for \nlanguages{} languages as available in
the public edition of the CorefUD~1.0 collection \cite{corefud2022lrec} and
follows the train/dev/test split of the collection, too.

\subsection{Data Resources}

Key features of the
original coreference resources harmonized under the CorefUD scheme are extracted from 
\citet{corefud2022lrec} into the following paragraphs; some of their quantitative properties are
summarized in Table~\ref{table:sizes}.

\begin{table*}[!ht]
  \begin{center}
  \begin{tabular}{@{}lrrrrr@{~~}rr@{}}\toprule 
\textbf{CorefUD dataset} &  \textbf{docs} &   \textbf{sents} &     \textbf{words} & \textbf{zeros} &  \textbf{entities} & \textbf{avg. len.} & \textbf{non-singletons} \\ \midrule
Catalan-AnCora           &  1550 &  16,678 &   546,665 &   6,377 &  69,239 & 1.6 &  62,416 \\
Czech-PCEDT              &  2312 &  49,208 & 1,155,755 &  43,054 &  52,743 & 3.4 & 178,376 \\
Czech-PDT                &  3165 &  49,428 &   834,721 &  32,617 &  78,880 & 2.5 & 169,545 \\
English-GUM              &   175 &   9,130 &   164,392 &      92 &  24,801 & 1.9 &  28,054 \\
English-ParCorFull       &    19 &     543 &    10,798 &       0 &     180 & 4.0 &     718 \\
French-Democrat          &   126 &  13,054 &   284,823 &       0 &  40,937 & 2.0 &  47,172 \\
German-ParCorFull        &    19 &     543 &    10,602 &       0 &     259 & 3.5 &     896 \\
German-PotsdamCC         &   176 &   2,238 &    33,222 &       0 &   3,752 & 1.4 &   2,519 \\
Hungarian-SzegedKoref    &   400 &   8,820 &   123,968 &   4,857 &   5,182 & 3.0 &  15,165 \\
Lithuanian-LCC           &   100 &   1,714 &    37,014 &       0 &   1,224 & 3.7 &   4,337 \\
Polish-PCC               &  1828 &  35,874 &   538,885 &     470 & 127,688 & 1.5 &  82,804 \\
Russian-RuCor            &   181 &   9,035 &   156,636 &       0 &   3,636 & 4.5 &  16,193 \\
Spanish-AnCora           &  1635 &  17,662 &   559,782 &   8,112 &  73,210 & 1.7 &  70,664 \\
\bottomrule\end{tabular}

  \caption{Data sizes in terms of the total number of documents, sentences, tokens, zeros (empty words), coreference entities, average entity length (in number of mentions) and the total number of non-singleton mentions. Train/dev/test splits of these datasets roughly follow 8/1/1 ratio. See~\citet{corefud2022lrec} for details. }
  \label{table:sizes}
  \end{center}
\end{table*}

\paragraph{Prague Dependency Treebank (Czech)}
(denoted as \cspdt{} for short in this paper) is a corpus of Czech newspaper
texts ($\sim$830K tokens) with manual multi-layer annotation
\cite{pdtconsolidated}. Coreference and bridging relations are annotated as
links on the deep syntactic layer. The links lead from the node of the
syntactic head of the anaphor to the node representing the syntactic head of
the antecedent and the whole subtrees of these nodes are considered to be
mention spans.

\paragraph{Prague Czech-English Dependency Treebank -- the Czech part}
(\cspcedt) is one side of the PCEDT parallel corpus
\cite{PCEDT2016coreference} consisting of more than 1M tokens. The annotation
of coreference-like phenomena is principally similar to the Prague Dependency
Treebank with some minor differences and no bridging annotation.

\paragraph{Georgetown University Multilayer Corpus (English)}
(\engum) \cite{GUMdocumentation} is a growing open source corpus of 12 written and spoken English genres ($\sim$180K tokens as of 2022). Next to UD syntax trees and discourse
parses, it exhaustively annotates all mentions, including nested,
named/non-named entities, singletons, and 10 entity classes and 6 information
status tags. It distinguishes 8 anaphoric links: pronominal anaphora and
cataphora, lexical and predicative coreference, apposition, discourse deixis,
split antecedents and bridging. 
For licence reasons, Reddit data is excluded from both the UD\_English-GUM and CorefUD~1.0 releases of GUM.

\paragraph{Polish Coreference Corpus}
(\plpcc) \cite{PCC2013,bookOgrodniczuk} is a corpus ($\sim$ 540K tokens) of
Polish nominal coreference built upon the National Corpus of Polish
\cite{przepiorkowski-etal-2008-towards}. Mentions
are annotated as linear spans, with additionally marked semantic heads. The
annotation includes identity coreference, quasi-identity relations and
non-identity close-to-coreference relations.

\paragraph{Democrat (French)}
(\frdemocrat) \cite{democrat} is a diachronic corpus of written French texts
from the 12th to the 21st century. The annotation focuses on nominal mentions
(pronouns and full NPs only) and includes information of definiteness and
syntactic type of mentions. Its conversion in CorefUD is based only on its
automatically parsed subset of texts from 19th-21st century \cite{dem1921-cr}
($\sim$280K tokens).

\paragraph{Russian Coreference Corpus}
(\rurucor) \cite{RuCorDialog} is a corpus of $\sim$150K tokens annotated with
anaphoric and coreferential relations between noun groups. Mentions are
annotated as linear spans, with additionally distinguished syntactic heads.
Only NPs which take part in coreference relations are considered and singletons
are not annotated.

\paragraph{ParCorFull (German and English)}
(\deparcorfull{} and \enparcorfull) is a parallel corpus of $\sim$160K tokens
annotated for coreference \cite{ParCorFullScheme}.  Mentions are NPs which
form part of pronoun-antecedent pairs, pronouns without antecedents or VPs if
they are antecedents of anaphoric NPs (discourse deixis). The annotation
includes identity coreference relations only. Due to license restrictions,
CorefUD contains only its WMT News section ($\sim$20K tokens).

\paragraph{AnCora: Multi-level Annotated Corpora for Catalan and Spanish}
(\caancora{} and \esancora) \cite{ancora,ancora-co} consist of very detailed
annotations of coreference (including zero anaphora, split antecedent,
discourse deixis, etc.). The corpora ($\sim$1M tokens) also contain
annotations of related phenomena such as argument structure, thematic roles,
semantic classes of verbs, named entities, denotative types of deverbal nouns
etc.

\paragraph{Potsdam Commentary Corpus (German)}
(\depotsdamcc) is a relatively small ($\sim$35K tokens) corpus of newspaper
articles \cite{bourgonje-stede-2020-potsdam} annotated for nominal and
pronominal identity coreference. Mentions are further classified into primary
(e.g.~pronouns, definite NPs, proper names), secondary (indefinite NPs,
clauses), and non-referring mentions. The corpus also contains gold
constituent syntax, information structure (including topic and focus, see
\citet{LuedelingRitzStedeEtAl2016}), and discourse parses.

\paragraph{Lithuanian Coreference Corpus}
(\ltlcc) \cite{LithuanianScheme} is a corpus of written texts, focusing on
political news ($\sim$35K tokens). Coreference annotation is link-based and
additional coreference information is divided into four levels that include
types of mentions, types of anaphoric relations, the direction of the
relation, and annotation of split antecedents.

\paragraph{SzegedKoref: Hungarian Coreference Corpus}
(\huszegedkoref) \cite{szegedkoref2018} is a corpus of written texts ($\sim$125K
tokens) selected from the Szeged Treebank. The treebank has manual
annotations at several linguistic layers such as deep phrase-structured
syntactic analysis, dependency syntax and morphology. Mentions are linear
spans without specially marked heads, the relations are classified into
anaphoric classes such as repetitions, synonyms, hypernyms, hyponyms etc.

\subsection{Annotation Details}

CorefUD collection is fully compliant with the CoNLL-U format,%
\footnote{%
\url{https://universaldependencies.org/format.html}
}
using the MISC column for annotation of coreference.
Besides coreference, also other anaphoric relations (e.g. bridging, split antecedents) are labeled in some CorefUD datasets.
Nevertheless, the shared task focuses only on coreference.
Therefore, the participants are asked to predict only the \textsf{Entity} attribute in the MISC column, namely the bracketing of mention spans (including possible discontinuities) and entity/cluster IDs assigning the mentions to entities.
They do not need to identify mention heads or fill other coreference-related features that can be found in CorefUD data.

Reconstructed zeros are an integral part of some of the CorefUD datasets.
CorefUD utilizes empty nodes in enhanced UD representation to mark them.
In the shared task data, we keep all annotated zeros and ask the participants to predict coreference also for them.
However, note that we decided not to strip off the empty nodes from the test data in the first edition of the shared task.
Although some datasets mark also non-anaphoric zeros, presence of an empty node may indicate its anaphoricity.
Its assignment to a cluster of other mentions still remains unknown, yet this makes the setup a bit unrealistic.
We find it a reasonable compromise between exploring insufficiently known area of zero anaphora in coreference resolution and making the shared task simple and accessible.

Apart from annotation of coreference and anaphora, CorefUD comprises also standard UD-like annotation of parts of speech, morphological features and dependency syntax.
With some exceptions, if the original resources contained manual annotation of morpho-syntax, it has been kept also in CorefUD.
Otherwise, it has been obtained automatically using UDPipe~2.0 \citep{udpipe2}.
Therefore, it must be noted that if a system takes advantage of this morpho-syntactic information, its performance on the datasets with manual morpho-syntax may be a bit overestimated, compared to real-world NLP scenarios in which manual annotations of morphology and syntax are usually not available.

\section{Evaluation Metrics}
\label{sec:evaluation}

Systems participating in the shared task are evaluated with the CorefUD
scorer.\footnote{\url{https://github.com/ufal/corefud-scorer}} The primary
evaluation score is the CoNLL $F_1$ score with singletons excluded
and using partial mention matching.
We also assess the shared task submissions by multiple
supplementary scores.

\paragraph{Official scorer}
We use our modification of the coreference scorer -- CorefUD scorer. It
is based on the Universal Anaphora (UA) scorer
\citep{UA-scorer-2022}\footnote{This in turn reimplements the official CoNLL-2012 scorer \citep{pradhan-etal-2014-scoring}.} reusing the implementations
of all generally used coreferential measures without any modification. This
guarantees that the measures are computed in exactly the same way. However,
our scorer is capable of processing the coreference annotation files in the
CorefUD~1.0 format.
Among other things, it allows evaluation of coreference for zeros.%
\footnote{%
Nonetheless, the current implementation is not able to handle a response document whose tokens are not completely identical to ones in the key document.
This holds also for empty nodes.
}
Moreover, it re-defines matching of key and response mentions in the way to be able to handle potentially discontinuous mentions, which are present in some CorefUD datasets.
Last but not least, we proposed and implemented the MM score to measure the accuracy of mention matching (see below).

\paragraph{Partial matching}
The CorefUD collection includes datasets (e.g. \cspdt{}) that do not
specify mention spans in their original annotations.
In these datasets, a mention is only specified by its head and loosely by a
dependency subtree rooted in this head. Also in other datasets, the exact
specification of mention boundaries may be difficult, for instance, if
mentions comprise embedded clauses, long detailed specifications, etc.
Therefore, authors of some datasets address this issue by defining a
syntactic or semantic head (single word) or a minimal span \citep[multiple words
possible, e.g. in ARRAU,][]{ARRAU2020}, i.e., a unit that carries the most important semantic information.

CorefUD specifies a mention head only syntactically. However, as it has been
shown in \citet{onehead-depling}, heads labeled within coreference annotation
most often correspond to heads defined by a dependency tree.

\begin{figure}
    \resizebox{\columnwidth}{!}{
    \centering
    \def\svgwidth{0.75\textwidth}
    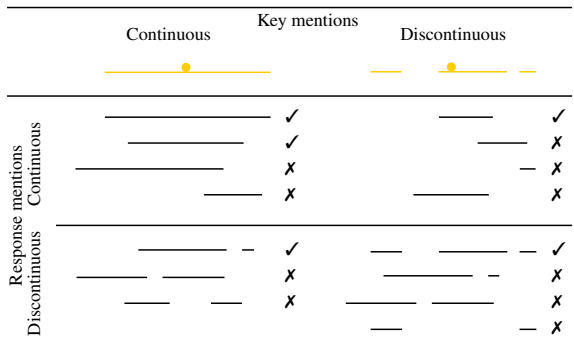
    }
    \caption{Examples of successful and unsuccessful partial mention matching 
    of key mentions (the yellow ones in the top; the mention head depicted by
    a small circle) by various response mentions.
    Showing cases of both continuous and discontinuous mentions.
    Recall the definition of partial match:
    A response mention matches a key mention if all its words are included in the key mention and one of them is the key head.
    }\label{fig:mention-matching}
\end{figure}

Availability of heads/minimal spans in key (i.e. gold reference) annotation allows for
\emph{partial mention matching} during the computation of any evaluation measure.
In the UA scorer, a response (i.e. predicted by a system) mention matches a key mention if the boundaries of the
response span lie within the key span and surround the key minimal span at
the same time.
In order to support evaluation of discontinuous mentions, we modified this
criterion using a set/subset relation. In the CorefUD scorer, a response
mention matches a key mention if all its words are included in the key
mention and one of them is the key head.
See Figure~\ref{fig:mention-matching} for examples of response mentions that
succeed or fail to match a key mention, depending on whether the mentions
are continuous or discontinuous.

\paragraph{Head matching}
The \emph{partial-match} approach to evaluation described above has two disadvantages.
First, it suffices for the systems to predict only heads instead of full mention spans.
For this reason, we report also the \emph{exact-match} version as a secondary measure.

Second, some authors may decide to post-process predictions of their systems
 by reducing the span of each mention to the head word only
 using one of the methods described below.
We can see in Table~\ref{tab:main-results} that five systems
 (\emph{straka*, berulasek and simple-rule-based})
 applied this post-processing and improved thus their results in terms of the primary metric.
However, this post-processing can be applied to any system,
 so we have decided to introduce it as another secondary metric called \emph{head-match}.
This way we can see what is the effect of such post-processing for systems
 which have not applied it.
The \emph{head-match} metric is even more benevolent than \emph{partial-match}
 because it does not penalize extra words added to the span
 as long as the head remains the same.

The shared task did not require 
 to predict the head in each mention.
However, the head can be predicted given the span and the provided dependency tree
 as the ``highest'' node.
We used Udapi block \texttt{corefud.MoveHead} for this purpose.\footnote{
 \url{https://github.com/udapi/udapi-python/blob/master/udapi/block/corefud/movehead.py}
 This block was used also for annotating the heads in the gold data.
}

The easiest post-processing method
 (chosen in all three \emph{straka*} submissions)
 is to reduce the span of each mention to the head.\footnote{
 With Udapi, it can be done using a command
  \texttt{udapy -s corefud.MoveHead
   util.Eval coref\_mention=\textquotesingle{}mention.words=[mention.head]\textquotesingle{}
   < in.conllu > out.conllu}.
}
However, the resulting CoNLL-U files may be invalid
 because two mentions may be assigned the same span.\footnote{
   For example in coordinations,
    the mention covering the whole coordination
    and the mention covering the first conjunct
    share the same head.
  It should be noted
   we did not require the submissions to pass the official UD validation tests
   (\texttt{validate.py -{}-level 2 -{}-coref}).
 }
One solution
 (chosen in the \emph{berulasek} submission)
 is to merge the entities of the two mentions
 which got assigned the same span.
In the \emph{head-match} solution, we chose a more conservative solution:
 if two spans share the same head,
 we reduce only the smaller span and keep the larger span intact.
We confirmed that differences between the three methods described in this paragraph
 according to the evaluation metrics are negligible
 because the cases of two mentions sharing the same head are rare.

\paragraph{Singletons}
The primary score is calculated excluding potential singletons, i.e.,
entities comprising only a single mention, in both key and response
coreference chains. We selected this option as the primary metric because
a majority of datasets in the CorefUD collection does not have singletons annotated.

\paragraph{Primary score}
As a primary evaluation metric, we employed the CoNLL $F_1$ score
\citep{CoNLL-MELA-score,pradhan-etal-2014-scoring}, which has been established as a standard for the evaluation of
coreference resolution. It is an unweighted average of $F_1$ scores of three
coreference measures: MUC \citep{MUC-score}, B$^3$ \citep{Bcubed-score}
and CEAF-e \citep{CEAF-score}, each adopting a different view on coreference
relations, namely link-based, mention-based and entity-based, respectively. A
single primary score providing a final ranking of participating submissions
is a macro-average over all datasets in the CorefUD test collection.

\paragraph{Supplementary scores}
In addition to the primary CoNLL $F_1$ score,
 we calculate three alternative versions of this metric:
 head-match, exact-match and with-singletons.

Besides the primary score and its three variants,
we also report the systems' performance in terms of two additional scores:
BLANC \citep{BLANC-score} and LEA \citep{LEA-score}.

In addition, we implement the MOR%
\footnote{It stands for Mention Overlap Ratio.}
score measuring to what extent key and response mentions match, no matter to which coreference entity they belong.
First, we find such one-to-one alignment $A(\mathcal{K}, \mathcal{R})$ between
the sets of all key mentions $\mathcal{K}$
 and all response mentions $\mathcal{R}$
 that maximizes the overall number of overlapping words within aligned mentions.
We then calculate the recall of mention overlap as a ratio of the total number of overlapping words in mentions and the overall size of all key  mentions (sum of its lengths):
\[MOR_{rec} = \frac{\sum_{(K,R) \in A(\mathcal{K}, \mathcal{R})}|K \cap R|}{\sum_{K \in \mathcal{K}}|K|}\]
Precision is calculated analogously using the set of all response mentions $\mathcal{R}$ in the denominator.
Note that position of the head in
mentions does not play a role in MOR score.
\todo{example?}

In order to show performance of the systems on zeros, we use an anaphor-decomposable score which is an application of the scoring schema introduced by \citet{tuggener-2014-coreference}.
For each zero mention other than the first one in the entity, we indicate a \emph{true positive (tp)} case if an overlap in at least one preceding mention is found between respective key and response entities.
\emph{Wrong linkage (wl)} is indicated if no such mention is found and \emph{False positive/negative (fp/fn)} case if the anaphoric response/key mention is not anaphoric (or it is the first mention of the entity) in the key/response document, respectively.
Having these counts aggregated, recall is calculated as $\frac{tp}{tp+wl+fn}$ and precision as $\frac{tp}{tp+wl+fp}$.

\section{Participating Systems}
\label{sec:systems}

\subsection{Baseline}
\label{sec:baseline}
The baseline system (\baseline{}\footnote{
 The baseline system was submitted to CodaLab under the name \emph{sidoj},
  but we rename it here to \baseline{} for clarity.
})
is based on the multilingual coreference resolution system presented by \citet{pravzak2021multilingual}.
The model uses multilingual BERT \cite{DBLP:journals/corr/abs-1810-04805} in the end-to-end setting.
In high-level terms, the model goes through all potential spans and maximizes the probability of gold antecedents for each span.
The same system is used for all the languages in the training dataset.

The simplified system adapted to CorefUD 1.0 is publicly available on GitHub\footnote{\url{https://github.com/ondfa/coref-multiling}} along with tagged dev data and its dev data results.

\subsection{System Comparison}

\begin{table*}[!ht]
\resizebox{\textwidth}{!}{
\centering
\begin{tabular}{llll}
\toprule \bf Team & \bf Submission                   & \bf Baseline based & \bf Approach  \\ \midrule
ÚFAL CorPipe      & straka                           & No                 & DL \\
                  & straka-single-multilingual-model & No                 & DL \\
                  & straka-only-single-treebank-data & No                 & DL \\
UWB               & ondfa                            & Yes                & DL \\
                  & \baseline                        & --                 & DL \\
Matouš Moravec    & Moravec                          & Yes -- files only  & rule-based postprocess of DL  \\
Barbora Dohnalová & berulasek                        & Yes -- files only  & rule-based postprocess of DL  \\
                  & simple-rule-based                & No                 & rules   \\
Karol Saputa      & k-sap                            & No                 & DL \\ \bottomrule
\end{tabular}
}
\caption{System comparison. The baseline solution, if involved, was either modified internally, or only its predictions were used and modified subsequently (``files only''). ``DL'' stands for a deep learning solution.
}
\label{tab:comparison}

\end{table*}

\begin{table*}
\resizebox{\textwidth}{!}{
\centering
\begin{tabular}{@{}lllllllll@{}}\toprule
\bf Team     & \bf Submission     & \bf Model                  & \bf SL & \bf Size & \bf Batch size & \bf Updates & \bf HParams \\ \midrule
ÚFAL CorPipe & straka             & google/rembert             & 512    & 614M & 8      & 960k    & 4          \\
             & straka-single\dots & google/rembert             & 512    & 614M & 8      & 960k    & 4          \\
             & straka-only\dots   & google/rembert             & 512    & 614M & 8      & 960k    & 4          \\
UWB          & ondfa              & xlm-roberta-large          & 512    & 600M & 1      & 800k    & 4          \\
             & \baseline          & multiling. BERT            & 512 & 220M & 1 & 800k & 0 \\
Karol Saputa & k-sap              & \makecell[tl]{allegro/herbert-\\base-cased} & 512      & 415M & Dynamic    & 27k     & $\sim$10  \\
\bottomrule
\end{tabular}
}
\caption{Machine Learning Parameters. SL means sequence length, Size is the number of trainable parameters in the models, Updates is the number of gradient updates during training and HParames shows the number of tuned hyper-parameters.}
\label{tab:ml_params}
\end{table*}

Table \ref{tab:comparison} shows the basic properties of all submitted systems for evaluation. The table is organized by individual teams. Some teams submitted more than one system. Roughly half of the systems exploited the provided baseline and the majority of the systems relied on machine learning.

Further details of the machine learning systems are  described in Table \ref{tab:ml_params}. The table indicates that all machine-learning systems rely on large pretrained models consisting of hundreds of millions of parameters. The ÚFAL CorPipe team and the UWB
team employ multilingual models. Karol Saputa utilizes a Polish model as he submitted results for Polish only. All teams who developed their deep-learning solution use the maximum sequence length of 512 sub-word tokens which equals the maximum allowed length of the employed models. Clearly, all the teams are aware of the necessity to model long dependencies in the coreference resolution task. The ÚFAL CorPipe trains on sentences and they put 8 samples in a batch. The UWB team works with documents and they put 1 document in a batch. Karol Saputa uses a dynamic batch to fill the buffer of 4\,000 subwords. The number of gradient updates is similar to the teams that train on all languages. Karol Saputa trains with a much smaller number of updates since he trains only on one corpus.

\subsection{Teams}
The descriptions below are based on the information provided
 by the respective participants in an online questionnaire.

\paragraph{ÚFAL CorPipe}
submitted three systems (for details see \cite{sharedtask-straka} in this volume). All are based on pre-trained masked language
models, either the RemBERT \cite{rembert} or the XLM-RoBERTa \cite{xlm-roberta} large models. Each sentence is
processed as an individual example. Additionally, the neighboring
sentences from the document are included as context -- the right context is limited
to 50 subwords, and the size of the left context is chosen so that the whole
input has 512 subwords. The model is trained jointly to perform two tasks --
mention span detection and coreference linking. The mention detection is
trained using a CRF sequence tagging scheme based on a generalization of BIO
encoding allowing overlapping mentions. Then, for each mention, it is decided
which of the preceding mentions is its antecedent (selecting the original
mention if there is no antecedent). To obtain a distribution over the
previous mentions, a query and a key are computed using a
nonlinear transformation, and then masked dot-product attention is utilized. Some experiments include \textit{corpus id} -- a special token at the beginning of a sample indicating the source corpus of the sample.

The \emph{straka} system is trained jointly on all training data in all languages. This strategy
exhibited a considerably better performance than training on individual
corpora separately. For each corpus, the optimal model and epoch is chosen according to
its development score. The \emph{straka-single-multilingual-model} system employs a single checkpoint of a single model, thus corresponding to a real deployment scenario. The chosen model
is based on Rembert, samples training data according to the logarithm of the respective corpus
size, and does not utilize the corpus id.
The \emph{straka-only-single-treebank-data} system uses an independent model for each corpus
with corresponding training data only. The model is based on Rembert,
and for each corpus the submitted predictions are from the epoch with the
best development performance.
All three submissions were post-processed by reducing mentions spans to the head
 (cf. Head matching in Section~\ref{sec:evaluation}).

\paragraph{UWB} submitted one system \emph{ondfa} which extends the baseline system (for details see \citet{sharedtask-prazak} in this volume). The system relies on combined datasets to employ cross-lingual training. The authors did not know the exact procedure to generate heads for mentions. Therefore, they attempted to learn the heads from the data. The system relies on XLM-Roberta large, which is a substantially bigger model than in the baseline.

\paragraph{Barbora Dohnalová}
submitted two systems, \emph{berulasek} and \emph{simple-rule-based},
 implemented as rule-based blocks in Udapi \citep{udapi}.\footnote{
  The \emph{simple-rule-based} system was originally called \emph{simple\_baseline} in CodaLab,
   but we renamed it here to prevent confusing it with the official baseline
   (described in Section~\ref{sec:baseline} and named \emph{sidoj} in CodaLab).
 }

\emph{berulasek} post-processes the baseline predictions
 by first reducing mention spans to the head
 (cf. Head matching in Section~\ref{sec:evaluation})
 and then adding all proper nouns (upos=PROPN) with the same lemma
 into the same entity cluster
 (potentially adding new mentions to existing entities).
The second step is applied only to cs, de, es, fr, and hu
 because it improved the results on the dev set only for these languages.

\emph{simple-rule-based} starts by
 linking each pronoun to the nearest previous noun of the same gender
 (as annotated in the provided CoNLL-U files)
 and then applies the ``\emph{berulasek}'' post-processing. 

The purpose of these two submissions was to show
 what results can be achieved with just a few lines of code
 and without using the training data.

\paragraph{Matouš Moravec} submitted one system \emph{moravec}.
The system is based on postprocessing existing coreference prediction using named entity information.
Specifically, the submission starts with baseline predictions, runs the NameTag web service\footnote{%
\url{http://lindat.mff.cuni.cz/services/nametag/api-reference.php}} \citep{strakova-etal-2019-neural}
on the underlying texts and applies the following three postprocessing rules using Udapi \citep{udapi}:
(1) changing coreference spans to spans of named entities,
(2) removing coreference links between different named entity types, and
(3) adding coreference links between named entities of the same type that have a high string similarity.
The author was not able to obtain any results that were better than the baseline for a whole dataset, although in some individual documents within these datasets coreference prediction was improved.

\paragraph{Karol Saputa} submitted one system \emph{k-sap} (for details see \cite{sharedtask-saputa} in this volume). It employs BERT-based antecedent scoring for possible spans based on representation of span start and end tokens. The submission employs the approach described by \citet{kirstain-etal-2021-coreference}.

\section{Results and Comparison}
\label{sec:results}

The \emph{straka} system by the ÚFAL CorPipe team is clearly the winner of the shared task.
It surpasses other systems not only in terms of the primary score (see the \emph{primary} column in Table~\ref{tab:main-results}) but consistently also in almost all coreference metrics, both in precision and recall (see Table~\ref{tab:secondary-metrics}).

Table~\ref{tab:all-langs} shows that systems submitted by the ÚFAL CorPipe team are dominant on the great majority of datasets.
They are outperformed only by the \emph{ondfa} system, namely on \deparcorfull{} and \huszegedkoref{} datasets.
Per-dataset evaluation also reveals that the last place of the \emph{k-sap} system in the overall ranking is unequivocally caused by ignoring all but the \plpcc{} dataset where it ranks 3rd.

In comparison to the baseline system, most systems
 ouperformed it by a relatively large margin.
The winning \emph{straka} system achieves over 12 points in the primary score, which is more than 20\% improvement over the baseline performance.
This is an extremely beneficial effect of the shared task, which may drive further development in multilingual coreference resolution.

Results unsurprisingly also confirm a doubtless dominance of machine learning approaches.
Although rule-based postprocessing has been employed by some teams (also encouraged by availability of the baseline predictions), its incorporation is either motivated by the nature of the primary score (\emph{straka*} systems) or it improves upon the baseline only marginally (the \emph{berulasek} system) or not at all (the \emph{moravec} system).

We observe almost the same picture in evaluation with singletons (see Table~\ref{tab:main-results}) -- the \emph{straka*} systems outperforming all the other systems.
Moreover, these submissions are the only ones that are positively affected by the inclusion of singletons.
It suggests that unlike other teams, ÚFAL CorPipe 
have optimized for singletons as well (confirmed by statistics on mentions and entities in Table~\ref{tab:stats-entities}).

Interestingly, no system outperformed the baseline in the exact-match evaluation (see the \emph{exact-match} column in Table~\ref{tab:main-results}).
Considerably low scores compared to the partial matching setting are apparently caused by the choice of partial matching as part of the primary score, which most of the teams optimized for.
Two teams (ÚFAL CorPipe and Barbora Dohnalová) even utilize the present dependency structure to reduce their mentions to heads only in post-processing (cf. Head matching in Section~\ref{sec:evaluation}).\footnote{
 It would be interesting to evaluate the ÚFAL CorPipe (\emph{straka*}) systems  before this post-processing,
  which slightly improves the primary metric (partial-match),
   but substantially worsens the exact-match.
}
The preference of most systems in underspecified mentions is confirmed by the head-match scores (Table~\ref{tab:main-results}), which are almost identical to the primary scores, and by MOR scores (see Table~\ref{tab:secondary-metrics}), reaching high precision but failing in recall.

\begin{table*}\centering

\begin{tabular}{@{}l r r@{~~}l r@{~~}l r@{~~}l @{}}\toprule
\bf system                        & \bf primary & \multicolumn{2}{c}{\bf head-match} & \multicolumn{2}{c}{\bf exact-match} & \multicolumn{2}{c}{\bf with-singletons} \\\midrule
straka               & \bf 70.72 & \bf 70.72 & (+0.00) &     33.18 & (-37.54) & \bf 72.98 & (+2.26)\\
straka-single\ldots  &     69.56 &     69.56 & (+0.00) &     33.06 & (-36.51) &     71.81 & (+2.25)\\
ondfa                &     67.64 &     68.51 & (+0.87) &     54.73 & (-12.91) &     58.06 & (-9.58)\\
straka-only\ldots    &     64.30 &     64.30 & (+0.00) &     32.28 & (-32.02) &     67.93 & (+3.63)\\
berulasek            &     59.72 &     59.72 & (+0.00) &     31.50 & (-28.22) &     50.84 & (-8.88)\\
\baseline            &     58.53 &     59.67 & (+1.13) & \bf 56.72 & (-1.82) &     49.69 & (-8.84)\\
moravec              &     55.05 &     56.35 & (+1.29) &     52.68 & (-2.37) &     46.79 & (-8.27)\\
simple-rule-based    &     18.14 &     18.14 & (+0.00) &     12.60 & (-5.54) &     17.13 & (-1.00)\\
k-sap                &      5.90 &      5.93 & (+0.03) &      5.84 & (-0.05) &      3.83 & (-2.07)\\
\bottomrule\end{tabular}

\caption{Main results: the CoNLL metric macro-averaged over all datasets.
The table shows the primary metric (partial-match, excluding singletons) and its three versions: head-match, exact-match and with-singletons.
The best score in each column is in bold.
}
\label{tab:main-results}
\end{table*}

\begin{table*}\centering
\begin{tabular}{@{}l cccccc @{}}\toprule
\bf system          & \bf MUC &\bf B$^3$ &\bf CEAF-e &\bf BLANC &\bf LEA &\bf MOR\\\midrule
straka              &  {\bf 74} /      76  / {\bf 74}  &  {\bf 67} / {\bf 72} / {\bf 68}  &  {\bf 71} / {\bf 70} / {\bf 70}  &  {\bf 63} /      70  / {\bf 65}  &  {\bf 63} / {\bf 69} / {\bf 65}  &       32  /      83  /      45  \\
straka-single\ldots &       72  /      76  /      73   &       65  /      72  /      67   &       67  /      70  /      68   &       61  / {\bf 71} /      64   &       62  /      68  /      64   &       32  /      84  /      45  \\
ondfa               &       69  / {\bf 76} /      72   &       61  /      71  /      65   &       62  /      69  /      65   &       59  /      69  /      63   &       58  /      67  /      62   &  {\bf 52} /      84  / {\bf 62} \\
straka-only\ldots   &       65  /      71  /      68   &       58  /      68  /      62   &       61  /      67  /      63   &       55  /      66  /      59   &       54  /      63  /      58   &       30  /      83  /      43  \\
berulasek           &       58  /      76  /      64   &       50  /      70  /      57   &       52  /      67  /      58   &       46  /      70  /      53   &       45  /      66  /      53   &       27  / {\bf 88} /      40  \\
\baseline           &       56  /      74  /      63   &       48  /      69  /      56   &       51  /      66  /      57   &       45  /      68  /      51   &       44  /      64  /      51   &       49  /      86  /      61  \\
moravec             &       53  /      70  /      60   &       45  /      65  /      52   &       50  /      59  /      53   &       41  /      59  /      46   &       41  /      60  /      48   &       49  /      81  /      60  \\
simple-rule-based   &       14  /      22  /      16   &       14  /      26  /      17   &       23  /      27  /      22   &       10  /      20  /      11   &        7  /      17  /       9   &       16  /      55  /      23  \\
k-sap               &        6  /       7  /       6   &        5  /       7  /       6   &        5  /       6  /       6   &        5  /       7  /       6   &        5  /       6  /       6   &        5  /       7  /       6  \\
\bottomrule\end{tabular}

\caption{Recall / Precision / F1 for individual secondary metrics.
All scores macro-averaged over all datasets.
Note that the high recall and F1 MOR scores of \textsc{ondfa}
 (relative to \textsc{straka*} systems)
 is caused by the fact that \textsc{ondfa} does not use any post-processing restricting mention spans to the head.
}
\label{tab:secondary-metrics}
\end{table*}

\newcommand*\rot{\rotatebox{90}}
\begin{table*}\centering
\small

\begin{tabular}{@{}l r@{~~~}r@{~~~}r@{~~~}r@{~~~}r@{~~~}r@{~~~}r@{~~~}r@{~~~}r@{~~~}r@{~~~}r@{~~~}r@{~~~}r@{}}\toprule
\bf system          & \rot\caancora & \rot\cspcedt & \rot\cspdt & \rot\deparcorfull & \rot\depotsdamcc & \rot\engum & \rot\enparcorfull & \rot\esancora & \rot\frdemocrat & \rot\huszegedkoref & \rot\ltlcc & \rot\plpcc & \rot\rurucor\\\midrule
straka              &    78.18 &\bf 78.59 &\bf 77.69 &    65.52 &    70.69 &    72.50 &\bf 39.00 &\bf 81.39 &\bf 65.27 &    63.15 &\bf 69.92 &    78.12 &\bf 79.34\\
straka-single\ldots &\bf 78.49 &    78.49 &    77.57 &    59.94 &\bf 71.11 &\bf 73.20 &    33.55 &    80.80 &    64.35 &    63.38 &    67.38 &\bf 78.32 &    77.74\\
ondfa               &    70.55 &    74.07 &    72.42 &\bf 73.90 &    68.68 &    68.31 &    31.90 &    72.32 &    61.39 &\bf 65.01 &    68.05 &    75.20 &    77.50\\
straka-only\ldots   &    76.34 &    77.87 &    76.76 &    36.50 &    56.65 &    70.66 &    23.48 &    78.78 &    64.94 &    62.94 &    61.32 &    73.36 &    76.26\\
berulasek           &    64.67 &    70.56 &    67.95 &    38.50 &    57.70 &    63.07 &    36.44 &    66.61 &    56.04 &    55.02 &    65.67 &    65.99 &    68.17\\
\baseline           &    63.74 &    70.00 &    67.27 &    33.75 &    55.44 &    62.59 &    36.44 &    65.99 &    55.55 &    52.35 &    64.81 &    65.34 &    67.66\\
moravec             &    58.25 &    68.19 &    64.71 &    31.86 &    52.84 &    59.15 &    36.44 &    62.01 &    54.87 &    52.00 &    59.49 &    63.40 &    52.49\\
simple-rule-based   &    15.58 &     5.51 &     9.48 &    29.81 &    19.41 &    21.99 &    11.37 &    16.64 &    21.74 &    17.00 &    27.53 &    15.69 &    24.06\\
k-sap               &     0.00 &     0.00 &     0.00 &     0.00 &     0.00 &     0.00 &     0.00 &     0.00 &     0.00 &     0.00 &     0.00 &    76.67 &     0.00\\
\bottomrule\end{tabular}

\caption{Results for individual languages in the primary metric (CoNLL).
}
\label{tab:all-langs}
\end{table*}

\begin{table*}\centering

\begin{tabular}{@{}l r@{~~~}r@{~~~}r@{~~~}r@{~~~}r@{~~~}r@{~~~}r@{~~~}r@{~~~}r@{~~~}r@{~~~}r@{~~~}r@{~~~}r@{}}\toprule
\bf system & \rot\caancora & \rot\cspdt & \rot\cspcedt & \rot\esancora & \rot\huszegedkoref & \rot\plpcc \\
\midrule
straka & {\bf 91} / 91 / {\bf 91} & {\bf 91} / {\bf 92} / {\bf 92} & 87 / {\bf 90} / {\bf 89} & {\bf 94} / {\bf 95} / {\bf 95} & 79 / 71 / 75 & 62 / 60 / 61 \\
straka-single\ldots & 91 / {\bf 92} / 91 & 91 / 92 / 92 & {\bf 88} / 90 / 89 & 94 / 95 / 95 & 76 / {\bf 76} / 76 & {\bf 79} / 83 / {\bf 81} \\
ondfa & 88 / 88 / 88 & 88 / 92 / 90 & 85 / 89 / 87 & 92 / 94 / 93 & {\bf 81} / 74 / {\bf 77} & 62 / 60 / 61 \\
straka-only\ldots & 89 / 88 / 88 & 90 / 92 / 91 & 87 / 89 / 88 & 92 / 92 / 92 & 74 / 70 / 72 & 71 / 63 / 67 \\
berulasek & 82 / 83 / 82 & 84 / 86 / 85 & 80 / 84 / 82 & 87 / 89 / 88 & 55 / 54 / 54 & 42 / 50 / 45 \\
\baseline & 82 / 82 / 82 & 84 / 86 / 85 & 80 / 83 / 82 & 87 / 88 / 87 & 52 / 51 / 52 & 42 / 50 / 45 \\
moravec & 81 / 82 / 82 & 84 / 85 / 84 & 80 / 83 / 81 & 87 / 88 / 87 & 52 / 51 / 52 & 42 / 50 / 45 \\
simple-rule-based & 0 / 0 / 0 & 0 / 0 / 0 & 0 / 0 / 0 & 0 / 0 / 0 & 0 / 0 / 0 & 0 / 0 / 0 \\
k-sap & 0 / 0 / 0 & 0 / 0 / 0 & 0 / 0 / 0 & 0 / 0 / 0 & 0 / 0 / 0 & 4 / {\bf 100} / 8 \\
\bottomrule\end{tabular}

\caption{Recall / Precision / F1 for anaphor-decomposable score of coreference resolution on zero anaphors across individual languages.
Only the datasets that contain anaphoric zeros are listed (\engum{} excluded as all zeros in its test set are non-anaphoric).
Note that these scores are directly comparable neither to the CoNLL score nor to the supplementary scores calculated with respect to whole entities in Table~\ref{tab:secondary-metrics}.
}
\label{tab:all-langs-zero}
\end{table*}


\subsection{Automatic analysis}

To the best of our knowledge, this is the first shared task on multilingual coreference resolution that includes zeros.
Therefore, Table~\ref{tab:all-langs-zero} focuses more on the performance with respect to zero anaphora (cf. Table~\ref{table:sizes} for proportion of all zeros in the data).
It shows anaphor-decomposable scores achieved by the systems on zeros across the datasets that comprise anaphoric zeros.
The best-performing systems surpass 90 F1 points for most of the languages.
Nevertheless, recall that the setup for zeros is slightly unrealistic as participants have been given the input documents with zeros (both anaphoric and non-anaphoric) already reconstructed.

We provide several additional tables in the appendices to shed more light
 on the differences between the submitted systems.
Table~\ref{tab:upos-entity} shows results factorized
 according the different part of speech tags in the mention heads.
Tables \ref{tab:stats-entities}--\ref{tab:stats-details}
 show various statistics on the entities and mentions
 in a concatenation of all the test sets.
Tables \ref{tab:stats-entities-pcedt}--\ref{tab:stats-details-pcedt}
 show the same statistics for \cspcedt{},
 which is the largest dataset in CorefUD~1.0.

\todo{promote the per anaphor type evaluation to the body}

\subsection{Manual analysis}

In addition to numerical scores, we also want to gain some insight into the
types of errors that individual systems do. Such error analysis is inevitably
incomplete, as we cannot manually check over 50,000 non-singleton mentions
from all the test datasets, times eight system submissions. Nevertheless,
here are some observations for illustration:

\subsubsection*{\baseline{}, \cspdt{}}
It often does not recognize a mention. For example, adjectives
derived from locations (\textit{ostravské} ``Ostrava-based'') tend
to be mentions in CorefUD, often nested ones
(\textit{ostravské firmy} ``Ostrava-based companies'') but the
system does not recognize them. 
It also fails to recognize many mentions that are regular noun phrases.

Once the system detects a mention, it often has the correct mention span, although there are some odd failures, too.
        
In case of a newspaper interview, first and second person
pronouns are recognized as mentions, coreference between
mentions of the same person is found correctly, but their
link to a person's name is easily misinterpreted.

\subsubsection*{\emph{straka}, \cspdt{}}
It detects some mentions that \baseline{} does
not see (e.g. \textit{ostravské}).


Linking names to first and second person pronouns is also a
problem, although the system got right one instance where the
baseline failed.

\subsubsection*{\baseline{}, \esancora{}}
There is an even more dramatic disproportion between number of mentions
found and those in the gold data.
This is probably caused by the large number of singletons in AnCora.

On the other hand, it correctly identified mentions
(including coreference) that were not annotated in the gold
data: M$_1$ = \textit{tanto China como Perú} ``China as well
as Peru'', M$_2$ = \textit{estas dos naciones} ``these two nations''.

Elsewhere, the coreference resolver got misled by similar
titles of two different people: \textit{el canciller peruano}
``the Peruvian secretary'' was linked to \textit{el canciller
chino} ``the Chinese secretary''.

\subsubsection*{\emph{straka}, \esancora{}}
Much more successful in identifying mentions; unlike the
baseline, it seems to be able to identify singletons.



Unlike the baseline, \emph{straka} did not recognize
\textit{tanto China como Perú} as a mention. It also did not
link the word \textit{China} from this expression to a
previous (singleton) instance of \textit{China}; but since
the same surprising annotation appears in the gold data, the
system scored here.

\section{Conclusions and Future Work}
\label{sec:conclusions}

This paper summarizes the outcomes of the Multilingual Coreference Resolution Shared Task held with the CRAC 2022 workshop. We hope that the presented shared task establishes a new state of the art
in multilingual coreference resolution.

Possible future editions of the shared task could be improved or extended
along the following directions:
\begin{itemize}
\item We will fix minor errors in CorefUD's harmonization procedure
    that have been identified during the shared task.
\item We would like to include additional datasets, especially for
    languages that have not been covered in CorefUD yet; about 20
    resources that have not been harmonized yet due to various reasons
    are listed in \citet{ourtechrep2021} (or have been harmonized, but
    cannot be distributed publicly because of license limitations).
\item We will try to find ways to include also coreference data from
    the OntoNotes project, which would be extremely valuable because of
    their size, quality, and popularity.
\item We will make the setup more realistic.
Firstly, we will delete empty nodes from the test data to be processed by participants' systems.
It also requires adjusting the scorer so that it can evaluate pairs of documents with different sets of empty nodes.
Secondly, we will replace the manual morpho-syntax annotation with the automatic one for the shared task.
\item We will consider introducing subtasks focused on other anaphoric
    relations than just identity coreference (see \citet{UA-scorer-2022}
    for a description of Universal Anaphora Scorer that is capable of
    evaluating also non-identity coreference relations);    for
    instance, some CorefUD datasets contain hand-annotated bridging
    relations already now.
\end{itemize}

\section*{Acknowledgements}

This work was supported by the Grant 20-16819X (LUSyD) of the Czech Science Foundation (GAČR); 
LM2018101 (LINDAT/CLARIAH-CZ) of the Ministry of Education, Youth, and Sports of the Czech Republic;
FW03010656 of the Technology Agency of the Czech Republic;
the European Regional Development Fund as a part of 2014--2020 Smart Growth Operational Programme, CLARIN --- Common Language Resources and Technology Infrastructure, project no. POIR.04.02.00-00C002/19; the project co-financed by the Polish Ministry of Education and Science under the agreement 2022/WK/09; 
and SGS-2022-016 Advanced methods of data processing and analysis. 
We thank all the participants of the shared task for participating
 and for providing brief descriptions of their systems. We also thank two anonymous reviewers and Jana Straková for very insightful and useful remarks.

\bibliography{anthology,custom}

\begin{thebibliography}{45}
\expandafter\ifx\csname natexlab\endcsname\relax\def\natexlab#1{#1}\fi

\bibitem[{Bagga and Baldwin(1998)}]{Bcubed-score}
Amit Bagga and Breck Baldwin. 1998.
\newblock Algorithms for scoring coreference chains.
\newblock In \emph{Proceedings of The First International Conference on
  Language Resources and Evaluation Workshop on Linguistics Coreference}, pages
  563--566.

\bibitem[{Bourgonje and Stede(2020)}]{bourgonje-stede-2020-potsdam}
Peter Bourgonje and Manfred Stede. 2020.
\newblock \href {https://www.aclweb.org/anthology/2020.lrec-1.133} {The
  {P}otsdam commentary corpus 2.2: Extending annotations for shallow discourse
  parsing}.
\newblock In \emph{Proceedings of the 12th Language Resources and Evaluation
  Conference}, pages 1061--1066, Marseille, France. European Language Resources
  Association.

\bibitem[{Buchholz and Marsi(2006)}]{buchholz2006conll}
Sabine Buchholz and Erwin Marsi. 2006.
\newblock {CoNLL-X} {S}hared {T}ask on {M}ultilingual {D}ependency {P}arsing.
\newblock In \emph{Proceedings of the Tenth Conference on Computational Natural
  Language Learning (CoNLL-X)}, pages 149--164.

\bibitem[{Chung et~al.(2020)Chung, F{\'{e}}vry, Tsai, Johnson, and
  Ruder}]{rembert}
Hyung~Won Chung, Thibault F{\'{e}}vry, Henry Tsai, Melvin Johnson, and
  Sebastian Ruder. 2020.
\newblock \href {http://arxiv.org/abs/2010.12821} {Rethinking embedding
  coupling in pre-trained language models}.
\newblock \emph{CoRR}, abs/2010.12821.

\bibitem[{Conneau et~al.(2019)Conneau, Khandelwal, Goyal, Chaudhary, Wenzek,
  Guzm{\'{a}}n, Grave, Ott, Zettlemoyer, and Stoyanov}]{xlm-roberta}
Alexis Conneau, Kartikay Khandelwal, Naman Goyal, Vishrav Chaudhary, Guillaume
  Wenzek, Francisco Guzm{\'{a}}n, Edouard Grave, Myle Ott, Luke Zettlemoyer,
  and Veselin Stoyanov. 2019.
\newblock \href {http://arxiv.org/abs/1911.02116} {Unsupervised cross-lingual
  representation learning at scale}.
\newblock \emph{CoRR}, abs/1911.02116.

\bibitem[{de~Marneffe et~al.(2021)de~Marneffe, Manning, Nivre, and Zeman}]{ud}
Marie-Catherine de~Marneffe, Christopher Manning, Joakim Nivre, and Daniel
  Zeman. 2021.
\newblock {Universal Dependencies}.
\newblock \emph{Computational Linguistics}, 47(2):255--308.

\bibitem[{Denis and Baldridge(2009)}]{CoNLL-MELA-score}
Pascal Denis and Jason Baldridge. 2009.
\newblock Global joint models for coreference resolution and named entity
  classification.
\newblock \emph{Proces. del Leng. Natural}.

\bibitem[{Devlin et~al.(2018)Devlin, Chang, Lee, and
  Toutanova}]{DBLP:journals/corr/abs-1810-04805}
Jacob Devlin, Ming{-}Wei Chang, Kenton Lee, and Kristina Toutanova. 2018.
\newblock \href {http://arxiv.org/abs/1810.04805} {{BERT:} pre-training of deep
  bidirectional transformers for language understanding}.
\newblock \emph{CoRR}, abs/1810.04805.

\bibitem[{Haji{\v{c}} et~al.(2020)Haji{\v{c}}, Bej{\v{c}}ek,
  Hlav{\'{a}}{\v{c}}ov{\'{a}}, Mikulov{\'{a}}, Straka, {\v{S}}těp{\'{a}}nek,
  and {\v{S}}t{\v{e}}p{\'{a}}nkov{\'{a}}}]{pdtconsolidated}
Jan Haji{\v{c}}, Eduard Bej{\v{c}}ek, Jaroslava Hlav{\'{a}}{\v{c}}ov{\'{a}},
  Marie Mikulov{\'{a}}, Milan Straka, Jan {\v{S}}těp{\'{a}}nek, and Barbora
  {\v{S}}t{\v{e}}p{\'{a}}nkov{\'{a}}. 2020.
\newblock \href {https://www.aclweb.org/anthology/2020.lrec-1.641.pdf} {{Prague
  Dependency Treebank - Consolidated 1.0}}.
\newblock In \emph{Proceedings of the 12th International Conference on Language
  Resources and Evaluation (LREC 2020)}, pages 5208--5218, Marseille, France.
  European Language Resources Association.

\bibitem[{Kirstain et~al.(2021)Kirstain, Ram, and
  Levy}]{kirstain-etal-2021-coreference}
Yuval Kirstain, Ori Ram, and Omer Levy. 2021.
\newblock \href {https://doi.org/10.18653/v1/2021.acl-short.3} {Coreference
  resolution without span representations}.
\newblock In \emph{Proceedings of the 59th Annual Meeting of the Association
  for Computational Linguistics and the 11th International Joint Conference on
  Natural Language Processing (Volume 2: Short Papers)}, pages 14--19, Online.
  Association for Computational Linguistics.

\bibitem[{Landragin(2021)}]{democrat}
Fr{\'e}d{\'e}ric Landragin. 2021.
\newblock \href {https://hal.archives-ouvertes.fr/hal-03474748} {{Le corpus
  Democrat et son exploitation. Pr{\'e}sentation}}.
\newblock \emph{{Langages}}, 224:11--24.

\bibitem[{Lapshinova-Koltunski et~al.(2018)Lapshinova-Koltunski, Hardmeier, and
  Krielke}]{ParCorFullScheme}
Ekaterina Lapshinova-Koltunski, Christian Hardmeier, and Pauline Krielke. 2018.
\newblock {ParCorFull: a Parallel Corpus Annotated with Full Coreference}.
\newblock In \emph{Proceedings of the Eleventh International Conference on
  Language Resources and Evaluation (LREC 2018)}, Miyazaki, Japan. European
  Language Resources Association (ELRA).

\bibitem[{L\"{u}deling et~al.(2016)L\"{u}deling, Ritz, Stede, and
  Zeldes}]{LuedelingRitzStedeEtAl2016}
Anke L\"{u}deling, Julia Ritz, Manfred Stede, and Amir Zeldes. 2016.
\newblock {Corpus Linguistics and Information Structure Research}.
\newblock In Caroline F\'{e}ry and Shinichiro Ichihara, editors, \emph{The
  Oxford Handbook of Information Structure}, pages 599--617. Oxford University
  Press.

\bibitem[{Luo(2005)}]{CEAF-score}
Xiaoqiang Luo. 2005.
\newblock \href {https://doi.org/10.3115/1220575.1220579} {On coreference
  resolution performance metrics}.
\newblock In \emph{Proceedings of the Conference on Human Language Technology
  and Empirical Methods in Natural Language Processing}, {HLT} '05, pages
  25--32. Association for Computational Linguistics.

\bibitem[{Moosavi and Strube(2016)}]{LEA-score}
Nafise~Sadat Moosavi and Michael Strube. 2016.
\newblock \href {https://doi.org/10.18653/v1/P16-1060} {Which coreference
  evaluation metric do you trust? a proposal for a link-based entity aware
  metric}.
\newblock In \emph{Proceedings of the 54th Annual Meeting of the Association
  for Computational Linguistics (Volume 1: Long Papers)}, pages 632--642,
  Berlin, Germany. Association for Computational Linguistics.

\bibitem[{Nedoluzhko et~al.(2016)Nedoluzhko, Nov{\'a}k, Cinkov{\'a},
  Mikulov{\'a}, and M{\'\i}rovsk{\'y}}]{PCEDT2016coreference}
Anna Nedoluzhko, Michal Nov{\'a}k, Silvie Cinkov{\'a}, Marie Mikulov{\'a}, and
  Ji{\v{r}}{\'\i} M{\'\i}rovsk{\'y}. 2016.
\newblock \href {https://www.aclweb.org/anthology/L16-1026} {Coreference in
  {P}rague {C}zech-{E}nglish {D}ependency {T}reebank}.
\newblock In \emph{Proceedings of the Tenth International Conference on
  Language Resources and Evaluation ({LREC}'16)}, pages 169--176,
  Portoro{\v{z}}, Slovenia. European Language Resources Association (ELRA).

\bibitem[{Nedoluzhko et~al.(2021{\natexlab{a}})Nedoluzhko, Nov{\'{a}}k, Popel,
  {\v{Z}}abokrtsk{\'{y}}, and Zeman}]{ourtechrep2021}
Anna Nedoluzhko, Michal Nov{\'{a}}k, Martin Popel, Zden{\v{e}}k
  {\v{Z}}abokrtsk{\'{y}}, and Daniel Zeman. 2021{\natexlab{a}}.
\newblock \href {https://ufal.mff.cuni.cz/techrep/tr66.pdf} {Coreference meets
  {Universal Dependencies} – a pilot experiment on harmonizing coreference
  datasets for 11 languages}.
\newblock Technical Report~66, {\'{U}}{FAL} {MFF} {UK}, Praha, Czechia.

\bibitem[{Nedoluzhko et~al.(2022)Nedoluzhko, Nov{\'a}k, Popel,
  {\v{Z}}abokrtsk{\`y}, Zeldes, and Zeman}]{corefud2022lrec}
Anna Nedoluzhko, Michal Nov{\'a}k, Martin Popel, Zdeněk {\v{Z}}abokrtsk{\`y},
  Amir Zeldes, and Daniel Zeman. 2022.
\newblock Corefud 1.0: Coreference meets universal dependencies.
\newblock In \emph{Proceedings of LREC}.

\bibitem[{Nedoluzhko et~al.(2021{\natexlab{b}})Nedoluzhko, Novák, Popel,
  Žabokrtský, and Zeman}]{onehead-depling}
Anna Nedoluzhko, Michal Novák, Martin Popel, Zdeněk Žabokrtský, and Daniel
  Zeman. 2021{\natexlab{b}}.
\newblock {Is one head enough? Mention heads in coreference annotations
  compared with UD-style heads}.
\newblock In \emph{Proceedings of the Sixth International Conference on
  Dependency Linguistics (Depling, SyntaxFest 2021)}, pages 101--114,
  Stroudsburg, PA, USA. Association for Computational Linguistics.

\bibitem[{Ogrodniczuk et~al.(2013)Ogrodniczuk, Glowińska, Kopeć, Savary, and
  Zawisławska}]{PCC2013}
Maciej Ogrodniczuk, Katarzyna Glowińska, Mateusz Kopeć, Agata Savary, and
  Magdalena Zawisławska. 2013.
\newblock \href {https://doi.org/10.1007/978-3-319-43808-5\_17} {Polish
  coreference corpus}.
\newblock In \emph{Human Language Technology. Challenges for Computer Science
  and Linguistics - 6th Language and Technology Conference, {LTC} 2013,
  Pozna{\'{n}}, Poland, December 7-9, 2013. Revised Selected Papers}, volume
  9561 of \emph{Lecture Notes in Computer Science}, pages 215--226. Springer.

\bibitem[{Ogrodniczuk et~al.(2015)Ogrodniczuk, Głowińska, Kopeć, Savary, and
  Zawisławska}]{bookOgrodniczuk}
Maciej Ogrodniczuk, Katarzyna Głowińska, Mateusz Kopeć, Agata Savary, and
  Magdalena Zawisławska. 2015.
\newblock \href {http://www.degruyter.com/view/product/428667}
  {\emph{Coreference in {P}olish: Annotation, Resolution and Evaluation}}.
\newblock Walter De Gruyter.

\bibitem[{Popel et~al.(2021)Popel, {\v{Z}}abokrtsk{\'y}, Nedoluzhko, Nov{\'a}k,
  and Zeman}]{2021findings}
Martin Popel, Zden{\v{e}}k {\v{Z}}abokrtsk{\'y}, Anna Nedoluzhko, Michal
  Nov{\'a}k, and Daniel Zeman. 2021.
\newblock \href {https://doi.org/10.18653/v1/2021.findings-emnlp.303} {Do {UD}
  trees match mention spans in coreference annotations?}
\newblock In \emph{Findings of the Association for Computational Linguistics:
  EMNLP 2021}, pages 3570--3576, Punta Cana, Dominican Republic. Association
  for Computational Linguistics.

\bibitem[{Popel et~al.(2017)Popel, Žabokrtský, and Vojtek}]{udapi}
Martin Popel, Zdeněk Žabokrtský, and Martin Vojtek. 2017.
\newblock {Udapi: Universal API for Universal Dependencies}.
\newblock In \emph{NoDaLiDa 2017 Workshop on Universal Dependencies}, pages
  96--101, Göteborg, Sweden. Göteborgs universitet.

\bibitem[{Pradhan et~al.(2014)Pradhan, Luo, Recasens, Hovy, Ng, and
  Strube}]{pradhan-etal-2014-scoring}
Sameer Pradhan, Xiaoqiang Luo, Marta Recasens, Eduard Hovy, Vincent Ng, and
  Michael Strube. 2014.
\newblock \href {https://doi.org/10.3115/v1/P14-2006} {Scoring coreference
  partitions of predicted mentions: A reference implementation}.
\newblock In \emph{Proceedings of the 52nd Annual Meeting of the Association
  for Computational Linguistics (Volume 2: Short Papers)}, pages 30--35,
  Baltimore, Maryland. Association for Computational Linguistics.

\bibitem[{Pradhan et~al.(2012)Pradhan, Moschitti, Xue, Uryupina, and
  Zhang}]{conll-2012}
Sameer Pradhan, Alessandro Moschitti, Nianwen Xue, Olga Uryupina, and Yuchen
  Zhang. 2012.
\newblock \href {https://www.aclweb.org/anthology/W12-4501} {{C}o{NLL}-2012
  shared task: Modeling multilingual unrestricted coreference in
  {O}nto{N}otes}.
\newblock In \emph{Joint Conference on {EMNLP} and {C}o{NLL} - Shared Task},
  pages 1--40, Jeju Island, Korea. Association for Computational Linguistics.

\bibitem[{Pra{\v{z}}{\'a}k et~al.(2021)Pra{\v{z}}{\'a}k, Konop{\'\i}k, and
  Sido}]{pravzak2021multilingual}
Ond{\v{r}}ej Pra{\v{z}}{\'a}k, Miloslav Konop{\'\i}k, and Jakub Sido. 2021.
\newblock Multilingual coreference resolution with harmonized annotations.
\newblock In \emph{Proceedings of the International Conference on Recent
  Advances in Natural Language Processing (RANLP 2021)}, pages 1119--1123.

\bibitem[{Pražák and Konopik(2022)}]{sharedtask-prazak}
Ondřej Pražák and Miloslav Konopik. 2022.
\newblock {End-to-end Multilingual Coreference Resolution with Mention Head
  Prediction}.
\newblock In \emph{Proceedings of the CRAC 2022 Shared Task on Multilingual
  Coreference Resolution}. Association for Computational Linguistics.

\bibitem[{Przepi{\'o}rkowski et~al.(2008)Przepi{\'o}rkowski, G{\'o}rski,
  Lewandowska-Tomaszyk, and {\L}azi{\'n}ski}]{przepiorkowski-etal-2008-towards}
Adam Przepi{\'o}rkowski, Rafa{\l}~L. G{\'o}rski, Barbara Lewandowska-Tomaszyk,
  and Marek {\L}azi{\'n}ski. 2008.
\newblock \href
  {http://www.lrec-conf.org/proceedings/lrec2008/pdf/211_paper.pdf} {Towards
  the {N}ational {C}orpus of {P}olish}.
\newblock In \emph{Proceedings of the Sixth International Conference on
  Language Resources and Evaluation ({LREC}'08)}, Marrakech, Morocco. European
  Language Resources Association (ELRA).

\bibitem[{Recasens and Hovy(2011)}]{BLANC-score}
Marta Recasens and Eduard~H. Hovy. 2011.
\newblock \href {https://doi.org/10.1017/S135132491000029X} {{BLANC}:
  Implementing the rand index for coreference evaluation}.
\newblock \emph{Natural Language Engineering}, 17(4):485--510.
\newblock Tex.bibsource= dblp computer science bibliography, https://dblp.org
  tex.biburl= https://dblp.org/rec/bib/journals/nle/{RecasensH}11.

\bibitem[{Recasens et~al.(2010)Recasens, M{\`a}rquez, Sapena, Mart{\'\i},
  Taul{\'e}, Hoste, Poesio, and Versley}]{recasens-etal-2010-semeval}
Marta Recasens, Llu{\'\i}s M{\`a}rquez, Emili Sapena, M.~Ant{\`o}nia
  Mart{\'\i}, Mariona Taul{\'e}, V{\'e}ronique Hoste, Massimo Poesio, and
  Yannick Versley. 2010.
\newblock \href {https://aclanthology.org/S10-1001} {{S}em{E}val-2010 task 1:
  Coreference resolution in multiple languages}.
\newblock In \emph{Proceedings of the 5th International Workshop on Semantic
  Evaluation}, pages 1--8, Uppsala, Sweden. Association for Computational
  Linguistics.

\bibitem[{Recasens and Mart\'{\i}(2010)}]{ancora-co}
Marta Recasens and M.~Ant\`{o}nia Mart\'{\i}. 2010.
\newblock \href {https://doi.org/10.1007/s10579-009-9108-x} {{AnCora-CO:
  Coreferentially Annotated Corpora for Spanish and Catalan}}.
\newblock \emph{Lang. Resour. Eval.}, 44(4):315–345.

\bibitem[{Saputa(2022)}]{sharedtask-saputa}
Karol Saputa. 2022.
\newblock {Coreference Resolution for Polish and Beyond: Description of the
  Herferencer System for the CRAC 2022 Shared Task on Multilingual Coreference
  Resolution}.
\newblock In \emph{Proceedings of the CRAC 2022 Shared Task on Multilingual
  Coreference Resolution}. Association for Computational Linguistics.

\bibitem[{Straka(2018)}]{udpipe2}
Milan Straka. 2018.
\newblock \href {https://doi.org/10.18653/v1/K18-2020} {{UDP}ipe 2.0 prototype
  at {C}o{NLL} 2018 {UD} shared task}.
\newblock In \emph{Proceedings of the {C}o{NLL} 2018 Shared Task: Multilingual
  Parsing from Raw Text to Universal Dependencies}, pages 197--207, Brussels,
  Belgium. Association for Computational Linguistics.

\bibitem[{Straka and Straková(2022)}]{sharedtask-straka}
Milan Straka and Jana Straková. 2022.
\newblock {ÚFAL CorPipe at CRAC 2022: Effectivity of Multilingual Models for
  Coreference Resolution}.
\newblock In \emph{Proceedings of the CRAC 2022 Shared Task on Multilingual
  Coreference Resolution}. Association for Computational Linguistics.

\bibitem[{Strakov{\'a} et~al.(2019)Strakov{\'a}, Straka, and
  Hajic}]{strakova-etal-2019-neural}
Jana Strakov{\'a}, Milan Straka, and Jan Hajic. 2019.
\newblock \href {https://doi.org/10.18653/v1/P19-1527} {Neural architectures
  for nested {NER} through linearization}.
\newblock In \emph{Proceedings of the 57th Annual Meeting of the Association
  for Computational Linguistics}, pages 5326--5331, Florence, Italy.
  Association for Computational Linguistics.

\bibitem[{Taul{\'e} et~al.(2008)Taul{\'e}, Mart{\'\i}, and Recasens}]{ancora}
Mariona Taul{\'e}, M.~Ant{\`o}nia Mart{\'\i}, and Marta Recasens. 2008.
\newblock \href
  {http://www.lrec-conf.org/proceedings/lrec2008/pdf/35_paper.pdf} {{A}n{C}ora:
  Multilevel annotated corpora for {C}atalan and {S}panish}.
\newblock In \emph{Proceedings of the Sixth International Conference on
  Language Resources and Evaluation ({LREC}'08)}, Marrakech, Morocco. European
  Language Resources Association (ELRA).

\bibitem[{Toldova et~al.(2014)Toldova, Roytberg, Ladygina, Vasilyeva,
  Azerkovich, Kurzukov, Sim, Gorshkov, Ivanova, Nedoluzhko, and
  Grishina}]{RuCorDialog}
Svetlana Toldova, Anna Roytberg, Alina Ladygina, Maria Vasilyeva, Ilya
  Azerkovich, Matvei Kurzukov, G.~Sim, D.V. Gorshkov, A.~Ivanova, Anna
  Nedoluzhko, and Yulia Grishina. 2014.
\newblock {Evaluating Anaphora and Coreference Resolution for Russian}.
\newblock In \emph{Komp'juternaja lingvistika i intellektual'nye tehnologii. Po
  materialam ezhegodnoj Mezhdunarodnoj konferencii Dialog}, pages 681--695.

\bibitem[{Tuggener(2014)}]{tuggener-2014-coreference}
Don Tuggener. 2014.
\newblock \href {https://doi.org/10.3115/v1/E14-4045} {Coreference resolution
  evaluation for higher level applications}.
\newblock In \emph{Proceedings of the 14th Conference of the {E}uropean Chapter
  of the Association for Computational Linguistics, volume 2: Short Papers},
  pages 231--235, Gothenburg, Sweden. Association for Computational
  Linguistics.

\bibitem[{Uryupina et~al.(2020)Uryupina, Artstein, Bristot, Cavicchio, Delogu,
  Rodriguez, and Poesio}]{ARRAU2020}
Olga Uryupina, Ron Artstein, Antonella Bristot, Federica Cavicchio, Francesca
  Delogu, Kepa~J. Rodriguez, and Massimo Poesio. 2020.
\newblock \href {https://doi.org/10.1017/S1351324919000056} {Annotating a broad
  range of anaphoric phenomena, in a variety of genres: the {ARRAU} {Corpus}}.
\newblock \emph{Natural Language Engineering}, 26(1):95--128.

\bibitem[{Vilain et~al.(1995)Vilain, Burger, Aberdeen, Connolly, and
  Hirschman}]{MUC-score}
Marc Vilain, John Burger, John Aberdeen, Dennis Connolly, and Lynette
  Hirschman. 1995.
\newblock \href {https://aclanthology.org/M95-1005} {A model-theoretic
  coreference scoring scheme}.
\newblock In \emph{Sixth Message Understanding Conference ({MUC}-6):
  Proceedings of a Conference Held in {C}olumbia, {M}aryland, November 6-8,
  1995}.

\bibitem[{Vincze et~al.(2018)Vincze, Heged{\H{u}}s, Sliz-Nagy, and
  Farkas}]{szegedkoref2018}
Veronika Vincze, Kl{\'a}ra Heged{\H{u}}s, Alex Sliz-Nagy, and Rich{\'a}rd
  Farkas. 2018.
\newblock \href {https://www.aclweb.org/anthology/L18-1061} {{S}zeged{K}oref: A
  {H}ungarian coreference corpus}.
\newblock In \emph{Proceedings of the Eleventh International Conference on
  Language Resources and Evaluation ({LREC} 2018)}, Miyazaki, Japan. European
  Language Resources Association (ELRA).

\bibitem[{Wilkens et~al.(2020)Wilkens, Oberle, Landragin, and
  Todirascu}]{dem1921-cr}
Rodrigo Wilkens, Bruno Oberle, Fr{\'e}d{\'e}ric Landragin, and Amalia
  Todirascu. 2020.
\newblock \href {https://www.aclweb.org/anthology/2020.lrec-1.10} {{F}rench
  coreference for spoken and written language}.
\newblock In \emph{Proceedings of the 12th Language Resources and Evaluation
  Conference}, pages 80--89, Marseille, France. European Language Resources
  Association.

\bibitem[{Yu et~al.(2022)Yu, Khosla, Moosavi, Paun, Pradhan, and
  Poesio}]{UA-scorer-2022}
Juntao Yu, Sopan Khosla, Nafise~Sadat Moosavi, Silviu Paun, Sameer Pradhan, and
  Massimo Poesio. 2022.
\newblock \href {https://aclanthology.org/2022.lrec-1.521} {The universal
  anaphora scorer}.
\newblock In \emph{Proceedings of the Language Resources and Evaluation
  Conference}, pages 4873--4883, Marseille, France. European Language Resources
  Association.

\bibitem[{Zeldes(2017)}]{GUMdocumentation}
Amir Zeldes. 2017.
\newblock \href {https://doi.org/10.1007/s10579-016-9343-x} {{The GUM Corpus:
  Creating Multilayer Resources in the Classroom}}.
\newblock \emph{Language Resources and Evaluation}, 51(3):581--612.

\bibitem[{{\v Z}itkus and Butkien\.{e}(2018)}]{LithuanianScheme}
Voldemaras {\v Z}itkus and Rita Butkien\.{e}. 2018.
\newblock \href {https://doi.org/10.1109/SNAMS.2018.8554892} {{Coreference
  Annotation Scheme and Corpus for {Lithuanian} Language}}.
\newblock In \emph{Fifth International Conference on Social Networks Analysis,
  Management and Security, {SNAMS} 2018, Valencia, Spain, October 15-18, 2018},
  pages 243--250. {IEEE}.

\end{thebibliography}
\bibliographystyle{acl_natbib}

\appendix
\def\MC#1#2{\multicolumn{#1}{c}{#2}}
\clearpage
\onecolumn
\section{Partial CoNLL results by head UPOS}
\label{sec:stats-upos}

\begin{table}[H]\centering
\begin{tabular}{@{}l r@{~~~}r@{~~~}r@{~~~}r@{~~~}r@{~~~}r@{~~~}r@{~~~}r@{}}\toprule
\bf system         & \bf NOUN  & \bf PRON  & \bf PROPN & \bf DET   & \bf ADJ   & \bf VERB  & \bf ADV   & \bf NUM  \\\midrule
straka             & \bf 68.71 & \bf 73.72 & \bf 72.29 &     66.58 &     47.71 & \bf 38.44 & \bf 49.85 & \bf 48.30 \\
straka-single\ldots&     67.17 &     73.25 &     70.35 &     62.65 & \bf 49.84 &     36.91 &     45.77 &     44.97 \\
ondfa              &     66.04 &     71.43 &     70.72 & \bf 69.01 &     39.67 &     25.47 &     38.51 &     33.52 \\
straka-only\ldots  &     61.46 &     67.08 &     63.89 &     60.60 &     41.38 &     30.71 &     35.70 &     39.55 \\
berulasek          &     56.43 &     61.55 &     59.47 &     48.91 &     32.74 &     18.37 &     23.67 &     31.02 \\
\baseline          &     55.24 &     60.44 &     58.23 &     48.65 &     30.43 &     18.29 &     23.44 &     29.87 \\
moravec            &     52.91 &     58.82 &     52.43 &     46.80 &     27.49 &     18.19 &     23.41 &     29.22 \\
simple-rule-based  &     10.22 &     18.27 &     17.78 &      6.32 &      2.96 &      3.31 &      1.58 &      4.97 \\
k-sap              &      5.74 &      5.80 &      5.99 &      5.84 &      4.72 &      5.77 &      4.08 &      5.98 \\
\bottomrule\end{tabular}
\caption{CoNLL F1 score evaluated only on entities with heads of a given UPOS.
In both the gold and prediction files we deleted some entities before running the evaluation.
We kept only entities with at least one mention with a given head UPOS (universal part of speech tag).
For the purpose of this analysis,
 if the head node had deprel=flat children,
 their UPOS tags were considered as well,
 so for example in ``Mr./NOUN Brown/PROPN''
 both NOUN and PROPN were taken as head UPOS,
 so the entity with this mention will be reported in both columns NOUN and PROPN.
Otherwise, the CoNLL F1 scores are the same as in the primary metric,
 i.e. an unweighted average over all datasets, partial-match, without singletons.
Note that when distinguishing entities into events and nominal entities,
 the VERB column can be considered as an approximation of the performance on events.
One of the limitations of this approach is that copula is not treated as head in the Universal Dependencies,
 so e.g. phrase \textit{She is nice} is not considered for the VERB column,
 but for the ADJ column (head of the phrase is \textit{nice}).
}
\label{tab:upos-entity}
\end{table}

\section{Statistics of the submitted systems on concatenation of all test sets}
\label{sec:stats-concat}

\begin{table}[H]\centering
\begin{tabular}{@{}l rrrr rrrrr@{}}\toprule
                    & \MC{4}{entities}                  & \MC{5}{distribution of lengths}      \\\cmidrule(lr){2-5}\cmidrule(l){6-10}
system              &   total & per 1k & \MC{2}{length} &     1 &     2 &     3 &     4 &   5+ \\\cmidrule(lr){4-5}
                    &   count &  words &    max &  avg. &  [\%] &  [\%] &  [\%] &  [\%] & [\%] \\\midrule
gold                &  41,001 &    104 &    509 &   2.2 &  54.9 &  23.5 &   9.2 &   4.3 &   8.0 \\
\baseline           &   4,541 &     11 &    217 &  11.2 &   0.0 &  33.1 &   8.6 &   6.0 &  52.3 \\
berulasek           &   4,583 &     12 &    242 &  11.1 &   0.4 &  32.8 &   8.9 &   6.1 &  51.8 \\
k-sap               &   1,744 &      4 &     41 &   4.0 &   0.1 &  50.1 &  18.8 &   8.6 &  22.4 \\
moravec             &   5,469 &     14 &    210 &  10.8 &   1.8 &  28.2 &   9.6 &   4.6 &  55.8 \\
ondfa               &   4,628 &     12 &    174 &  11.7 &   0.0 &  31.6 &   9.5 &   5.4 &  53.5 \\
simple-rule-based   &   1,729 &      4 &    149 &  16.3 &   0.0 &   4.5 &   1.3 &   7.8 &  86.5 \\
straka              &  12,669 &     32 &    200 &   7.1 &  27.1 &   4.5 &   3.6 &   6.8 &  58.0 \\
straka-only\ldots   &  12,552 &     32 &    338 &   7.2 &  25.5 &   4.4 &   4.1 &   7.3 &  58.7 \\
straka-single\ldots &  12,669 &     32 &    243 &   7.1 &  26.2 &   4.4 &   4.0 &   6.9 &  58.5 \\
\bottomrule\end{tabular}
\caption{Statistics on coreference entities. The total number of entities and the average number of
entities per 1000 tokens in the running text. The maximum and average entity ``length'', i.e., number
of mentions in the entity. Distribution of entity lengths (singletons have length = 1).
}
\label{tab:stats-entities}
\end{table}

\begin{table}[H]\centering
\begin{tabular}{@{}l rrrr rrrrrr@{}}\toprule
                    & \MC{4}{mentions}                  & \MC{6}{distribution of lengths}              \\\cmidrule(lr){2-5}\cmidrule(l){6-11}
system              &   total & per 1k & \MC{2}{length} &     0 &     1 &     2 &     3 &     4 &   5+ \\\cmidrule(lr){4-5}
                    &   count &  words &    max &  avg. &  [\%] &  [\%] &  [\%] &  [\%] &  [\%] & [\%] \\\midrule
gold                &  69,406 &    175 &    104 &   3.3 &  10.2 &  39.4 &  19.6 &   8.5 &   4.4 &  17.9 \\
\baseline           &  50,783 &    128 &     26 &   2.2 &  13.3 &  46.3 &  19.1 &   7.3 &   3.4 &  10.7 \\
berulasek           &  50,935 &    129 &      1 &   0.9 &  13.4 &  86.6 &   0.0 &   0.0 &   0.0 &   0.0 \\
k-sap               &   6,941 &     18 &     29 &   1.6 &   0.0 &  75.1 &  14.1 &   4.1 &   2.0 &   4.7 \\
moravec             &  58,883 &    149 &     26 &   2.1 &  11.5 &  50.2 &  18.5 &   7.2 &   3.2 &   9.5 \\
ondfa               &  54,018 &    137 &     30 &   1.7 &  12.5 &  65.8 &   9.6 &   3.8 &   1.9 &   6.4 \\
simple-rule-based   &  28,130 &     71 &      1 &   1.0 &   0.0 & 100.0 &   0.0 &   0.0 &   0.0 &   0.0 \\
straka              &  86,412 &    218 &      1 &   0.9 &   8.4 &  91.6 &   0.0 &   0.0 &   0.0 &   0.0 \\
straka-only\ldots   &  87,059 &    220 &      1 &   0.9 &   8.4 &  91.6 &   0.0 &   0.0 &   0.0 &   0.0 \\
straka-single\ldots &  86,689 &    219 &      1 &   0.9 &   8.4 &  91.6 &   0.0 &   0.0 &   0.0 &   0.0 \\
\bottomrule\end{tabular}

\caption{Statistics on non-singleton mentions.
The total number of mentions and the average number of
mentions per 1000 words of running text. The maximum and average mention length, i.e., number of nonempty nodes in the mention. Distribution of mention lengths (zeros have length = 0).
}
\label{tab:stats-mentions-nonsingleton}
\end{table}

\begin{table}[H]\centering
\begin{tabular}{@{}l @{}r@{~}r@{~}r @{~}r@{~}r@{~}r@{~}r@{~}r@{~}r@{~}r@{~}r@{~}r@{}}\toprule
                    & \MC{3}{mention type [\%]}    & \MC{9}{distribution of head UPOS [\%]}      \\\cmidrule(lr){2-4}\cmidrule(l){5-13}
system              & w/empty & w/gap & non-tree
                                             &  NOUN &  PRON & PROPN &   DET &   ADJ &  VERB &   ADV &   NUM & other \\\midrule
gold                &   9.9 &   0.7 &   2.6 &  52.6 &  17.9 &  13.9 &   5.4 &   2.5 &   3.5 &   1.0 &   1.1 &   2.0 \\
\baseline           &  15.0 &   0.0 &   2.1 &  38.7 &  28.6 &  14.0 &   8.4 &   2.6 &   3.9 &   1.1 &   0.3 &   2.3 \\
berulasek           &  13.4 &   0.0 &   0.0 &  38.2 &  28.5 &  14.7 &   8.4 &   2.6 &   3.8 &   1.1 &   0.3 &   2.2 \\
k-sap               &   0.2 &   0.0 &   1.5 &  39.9 &  14.1 &  13.3 &   3.0 &   1.2 &  19.5 &   0.5 &   0.1 &   8.4 \\
moravec             &  12.9 &   0.0 &   2.4 &  35.0 &  24.6 &  21.7 &   7.7 &   2.3 &   3.5 &   1.0 &   0.4 &   3.9 \\
ondfa               &  13.3 &   0.0 &   1.4 &  40.7 &  27.6 &  13.6 &   8.1 &   2.6 &   3.6 &   1.2 &   0.4 &   2.3 \\
simple-rule-based   &   0.0 &   0.0 &   0.0 &  15.6 &  62.6 &  21.8 &   0.0 &   0.0 &   0.0 &   0.0 &   0.0 &   0.0 \\
straka              &   8.1 &   0.0 &   0.0 &  52.4 &  18.4 &  13.9 &   5.6 &   2.2 &   3.5 &   0.9 &   1.1 &   2.1 \\
straka-only\ldots   &   8.1 &   0.0 &   0.0 &  52.0 &  18.3 &  14.0 &   5.5 &   2.3 &   3.8 &   0.9 &   1.0 &   2.2 \\
straka-single\ldots &   8.1 &   0.0 &   0.0 &  52.4 &  18.3 &  14.1 &   5.6 &   2.2 &   3.5 &   0.8 &   1.0 &   2.1 \\
\bottomrule\end{tabular}
\caption{Detailed statistics on mentions.
The left part of the table shows percentage of:
 mentions with at least one empty node (w/empty);
 mentions with at least one gap, i.e. discontinuous mentions (w/gap);
 and non-treelet mentions, i.e. mentions not forming a connected subgraph in the dependency tree (non-tree).
Note that these three types of mentions may be overlapping.
The right part of the table shows distribution of mentions
 based on the universal part-of-speech tag (UPOS) of the head word.
Note that the participants were not required to predict the head,
 so we used Udapi block \texttt{corefud.MoveHead} on all submissions
 for the purpose of these statistics.
}
\label{tab:stats-details}
\end{table}

\section{Statistics of the submitted systems on \cspcedt{}}
\label{sec:stats-pcedt}

\begin{table}[H]\centering
\begin{tabular}{@{}l rrrr rrrrr@{}}\toprule
                    & \MC{4}{entities}                  & \MC{5}{distribution of lengths}      \\\cmidrule(lr){2-5}\cmidrule(l){6-10}
system              &   total & per 1k & \MC{2}{length} &     1 &     2 &     3 &     4 &   5+ \\\cmidrule(lr){4-5}
                    &   count &  words &    max &  avg. &  [\%] &  [\%] &  [\%] &  [\%] & [\%] \\\midrule
gold                &   2,533 &     45 &     89 &   3.3 &   1.8 &  63.7 &  14.8 &   6.4 &  13.3 \\
\baseline           &   2,048 &     37 &     78 &   3.5 &   0.0 &  62.1 &  16.5 &   6.1 &  15.4 \\
berulasek           &   2,062 &     37 &     80 &   3.5 &   0.7 &  62.2 &  15.8 &   6.0 &  15.3 \\
moravec             &   2,284 &     41 &     77 &   3.6 &   2.1 &  55.8 &  18.3 &   6.8 &  16.9 \\
ondfa               &   2,136 &     38 &     74 &   3.5 &   0.0 &  61.9 &  16.1 &   6.3 &  15.7 \\
simple-rule-based   &     271 &      5 &     57 &   6.1 &   0.0 &  46.1 &  14.4 &  11.1 &  28.4 \\
straka              &   2,770 &     49 &     81 &   3.0 &  16.4 &  50.1 &  15.2 &   6.4 &  11.9 \\
straka-only\ldots   &   2,741 &     49 &     80 &   3.0 &  16.9 &  48.9 &  15.0 &   6.8 &  12.4 \\
straka-single\ldots &   2,773 &     49 &     82 &   3.0 &  18.1 &  48.6 &  15.3 &   6.1 &  11.9 \\
\bottomrule\end{tabular}
\caption{Statistics on coreference entities in \cspcedt{}.
}
\label{tab:stats-entities-pcedt}
\end{table}

\begin{table}[H]\centering
\begin{tabular}{@{}l rrrr rrrrrr@{}}\toprule
                    & \MC{4}{mentions}                  & \MC{6}{distribution of lengths}              \\\cmidrule(lr){2-5}\cmidrule(l){6-11}
system              &   total & per 1k & \MC{2}{length} &     0 &     1 &     2 &     3 &     4 &   5+ \\\cmidrule(lr){4-5}
                    &   count &  words &    max &  avg. &  [\%] &  [\%] &  [\%] &  [\%] &  [\%] & [\%] \\\midrule
gold                &   8,365 &    149 &     61 &   3.6 &  22.6 &  26.9 &  17.4 &   8.6 &   3.9 &  20.6 \\
\baseline           &   7,258 &    129 &     22 &   2.5 &  24.6 &  28.2 &  18.7 &   9.0 &   4.1 &  15.4 \\
berulasek           &   7,262 &    130 &      1 &   0.8 &  24.9 &  75.1 &   0.0 &   0.0 &   0.0 &   0.0 \\
moravec             &   8,228 &    147 &     22 &   2.4 &  21.7 &  31.7 &  19.2 &   9.2 &   4.1 &  14.1 \\
ondfa               &   7,527 &    134 &     21 &   2.7 &  23.4 &  27.4 &  18.3 &   9.0 &   4.5 &  17.3 \\
simple-rule-based   &   1,640 &     29 &      1 &   1.0 &   0.0 & 100.0 &   0.0 &   0.0 &   0.0 &   0.0 \\
straka              &   7,890 &    141 &      1 &   0.8 &  24.0 &  76.0 &   0.0 &   0.0 &   0.0 &   0.0 \\
straka-only\ldots   &   7,888 &    141 &      1 &   0.8 &  24.1 &  75.9 &   0.0 &   0.0 &   0.0 &   0.0 \\
straka-single\ldots &   7,831 &    140 &      1 &   0.8 &  24.1 &  75.9 &   0.0 &   0.0 &   0.0 &   0.0 \\
\bottomrule\end{tabular}
\caption{Statistics on non-singleton mentions in \cspcedt{}.
}
\label{tab:stats-mentions-nonsingleton-pcedt}
\end{table}


\begin{table}[H]\centering
\begin{tabular}{@{}l @{}r@{~}r@{~}r @{~}r@{~}r@{~}r@{~}r@{~}r@{~}r@{~}r@{~}r@{~}r@{}}\toprule
                    & \MC{3}{mention type [\%]}    & \MC{9}{distribution of head UPOS [\%]}      \\\cmidrule(lr){2-4}\cmidrule(l){5-13}
system              & w/empty & w/gap & non-tree
                                             &  NOUN &  PRON & PROPN &   DET &   ADJ &  VERB &   ADV &   NUM & other \\\midrule
gold                &  29.2 &   1.2 &   4.5 &  44.9 &  28.0 &   6.4 &  12.4 &   0.9 &   2.7 &   1.5 &   0.7 &   2.6 \\
\baseline           &  28.3 &   0.0 &   3.8 &  45.1 &  30.2 &   6.4 &  12.2 &   0.6 &   1.5 &   1.3 &   0.7 &   2.0 \\
berulasek           &  24.9 &   0.0 &   0.0 &  44.6 &  30.1 &   7.2 &  12.2 &   0.5 &   1.5 &   1.3 &   0.7 &   1.9 \\
moravec             &  24.8 &   0.0 &   3.8 &  41.5 &  26.5 &  12.4 &  10.7 &   0.6 &   1.3 &   1.1 &   0.7 &   5.2 \\
ondfa               &  27.5 &   0.0 &   3.5 &  45.3 &  29.0 &   6.1 &  12.7 &   0.7 &   2.0 &   1.4 &   0.6 &   2.3 \\
simple-rule-based   &   0.0 &   0.0 &   0.0 &   3.4 &  78.2 &  18.4 &   0.0 &   0.0 &   0.0 &   0.0 &   0.0 &   0.0 \\
straka              &  23.3 &   0.0 &   0.0 &  45.0 &  28.1 &   5.9 &  12.7 &   0.8 &   2.7 &   1.3 &   0.7 &   2.8 \\
straka-only\ldots   &  23.2 &   0.0 &   0.0 &  44.9 &  28.2 &   6.1 &  12.5 &   1.0 &   2.8 &   1.3 &   0.6 &   2.7 \\
straka-single\ldots &  23.3 &   0.0 &   0.0 &  45.0 &  28.2 &   6.0 &  12.7 &   0.8 &   2.6 &   1.3 &   0.6 &   2.8 \\
\bottomrule\end{tabular}
\caption{Detailed statistics on mentions in \cspcedt.}
\label{tab:stats-details-pcedt}
\end{table}

\end{document}